\documentclass[10pt,twocolumn,letterpaper]{article}

\usepackage[pagenumbers]{cvpr} 

\definecolor{cvprblue}{rgb}{0.21,0.49,0.74}
\usepackage[pagebackref,breaklinks,colorlinks,allcolors=cvprblue]{hyperref}
\usepackage{booktabs}
\usepackage{array}
\usepackage{caption}
\usepackage{subcaption}
\usepackage{graphicx}
\usepackage{booktabs}
\usepackage[table]{xcolor}
\usepackage{subcaption}
\usepackage{amsmath}
\usepackage{colortbl}
\usepackage{xcolor}
\usepackage{booktabs}
\usepackage{multirow}
\usepackage{graphicx}
\usepackage{arydshln}
\definecolor{t2ibg}{gray}{0.92}
\usepackage{algorithm}
\usepackage{algorithmic}

\def\paperID{954} 
\def\confName{CVPR}
\def\confYear{2026}

\title{VOSR: A Vision-Only Generative Model for Image Super-Resolution}

\author{
Rongyuan Wu$^{1,2*}$
\quad
Lingchen Sun$^{1,2*}$
\quad
Zhengqiang Zhang$^{1,2}$
\quad
Xiangtao Kong$^{1,2}$\\
Jixin Zhao$^{1,2}$
\quad
Shihao Wang$^{1}$
\quad
Lei Zhang$^{1,2\dagger}$\\[0.5em]
$^{1}$The Hong Kong Polytechnic University
\quad
$^{2}$OPPO Research Institute
}

\begin{document}
\maketitle\begingroup
\renewcommand\thefootnote{}
\footnotetext{\hspace*{-1.5em}* Equal contribution. \hspace{0.1em} $\dagger$ Corresponding author. This research is supported by the PolyU-OPPO Joint Innovative Research Center.}
\endgroup
\begin{abstract}

Most of the recent generative image super-resolution (SR) methods rely on adapting large text-to-image (T2I) diffusion models pretrained on web-scale text-image data. While effective, this paradigm starts from a generic T2I generator, despite that SR is fundamentally a low-resolution (LR) input-conditioned image restoration task. In this work, we investigate whether an SR model trained purely on visual data can rival T2I-based ones. To this end, we propose \textbf{VOSR}, a \textbf{V}ision-\textbf{O}nly generative framework for \textbf{SR}. We first extract semantically rich and spatially grounded features from the LR input using a pretrained vision encoder as visual semantic guidance. We then revisit classifier-free guidance for training generative models and show that the standard unconditional branch is ill-suited to restoration models trained from scratch. We therefore replace it with a restoration-oriented guidance strategy that preserves weak LR anchors. Built upon these designs, we first train a multi-step VOSR model from scratch and then distill it into a one-step model for efficient inference. VOSR requires less than one-tenth of the training cost of representative T2I-based SR methods, yet in both multi-step and one-step settings, it achieves competitive or even better perceptual quality and efficiency, while producing more faithful structures with fewer hallucinations on both synthetic and real-world benchmarks. Our results, for the first time, show that high-quality generative SR can be achieved without multimodal pretraining. The code and models can be found at \href{https://github.com/cswry/VOSR}{https://github.com/cswry/VOSR}.
\end{abstract}

\begin{figure*}[t]
  \centering
  \includegraphics[width=\linewidth]{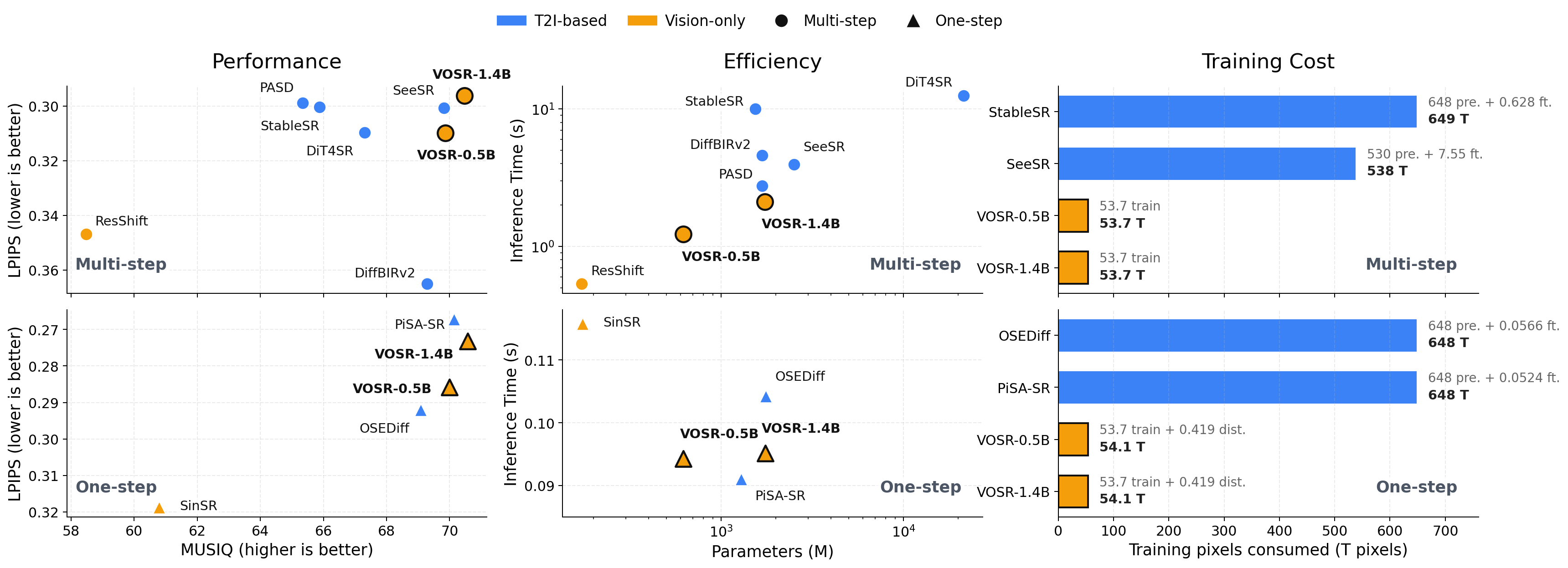}
  \vspace{-2mm}
  \caption{
    Comparison of VOSR with existing generative SR methods in terms of \textbf{performance}, \textbf{efficiency}, and \textbf{training cost}. Blue/orange colors denote T2I-based and vision-only methods, circles/triangles denote multi-step and one-step models. Performance is measured on RealSR \cite{realsr}, and efficiency is measured at $512\times512$ resolution using official repositories. VOSR achieves competitive or better perceptual quality than many T2I-based SR methods in both multi-step and one-step settings, while clearly outperforming prior vision-only methods. Its multi-step variant is substantially more efficient than existing T2I-based methods, and its one-step variant remains comparable to recent one-step T2I systems. Measured by the total number of training pixels consumed, VOSR requires only about one-tenth of the training cost of representative T2I-based SR methods. For fairness, we count only the pretraining cost of the core diffusion modules; reused components such as the VAE and semantic encoders are excluded.
  }
  \vspace{-2mm}
  \label{fig:overview}
\end{figure*}

\vspace{-3mm}
\section{Introduction}
\label{sec:intro}

Image super-resolution (SR) is a fundamental problem in low-level vision, aiming to reconstruct a high-resolution (HR) image from its degraded low-resolution (LR) observation. Unlike the free-form image synthesis problem, SR is an input-conditioned restoration problem: the reconstructed HR image should be perceptually both realistic and faithful to the LR input. Early deep learning based SR methods~\cite{dong2014learning, rcan, liang2021swinir, zhang2022efficient, chen2023activating} mainly optimize pixel-wise objectives such as $\ell_1$ loss, which improve distortion measures but often produce over-smoothed textures. GAN-based methods~\cite{srgan, wang2018esrgan, wang2021real} improve perceptual sharpness, but often suffer from unstable adversarial optimization and visible artifacts. More recently, diffusion models~\cite{ho2020denoising, song2020score} have emerged as a powerful alternative to GAN, combining stable training with strong generative capacity. Early vision-only diffusion SR methods, such as SR3~\cite{sr3}, SRDiff~\cite{li2022srdiff} and ResShift~\cite{yue2023resshift}, have demonstrated promising detail synthesis performance, but struggle in challenging real-world scenarios where the LR inputs show semantic ambiguity and complex degradations.

A recent line of research addresses this limitation by adapting pre-trained text-to-image (T2I) diffusion models, such as Stable Diffusion~\cite{sd, podell2023sdxl, esser2024scaling}, for SR~\cite{wang2023exploiting, wu2024seesr, sun2023improving, yang2023pixel, lin2024diffbir, chen2025faithdiff, duan2025dit4sr}. 
In general, these methods start with a generic T2I generator and then enforce compliance with the LR input through prompts, adapters, or other control mechanisms. Although effective, these methods adapt a generic T2I generator to process the LR input, rather than train a restoration model directly for detail generation. For SR, however, being perceptually realistic alone is insufficient; the restored details must also be faithful to the LR observation. Unfortunately, the multi-modal pre-training nature of T2I models introduces semantics through text or text-aligned representations \cite{wu2024seesr}, increasing the risk of detail hallucination. Even when derived from the LR input itself, such cues are still injected through a text-conditioned generative path, which is often spatially coarse and weakly grounded. These observations motivate us to raise a fundamental question: \textit{can a purely vision-based generative model, trained directly for restoration without relying on multimodal pre-training, rival T2I-based SR models?}

In this work, we answer this question by proposing \textbf{VOSR}, a \textbf{V}ision-\textbf{O}nly generative model for image \textbf{S}uper-\textbf{R}esolution. Here, \textit{vision-only} means that the model is trained only on visual data, including synthesized LR-HR pairs and auxiliary features from vision encoders, without multimodal text-image supervision. Rather than imitating the T2I training paradigm, VOSR follows a native restoration-oriented design. First, we introduce \textit{vision semantic condition}, which extracts semantically rich features directly in the visual domain from the LR image using pretrained vision encoders such as DINO~\cite{oquab2023dinov2, simeoni2025dinov3}. Unlike text-aligned semantics, these features are tightly grounded to the input and better suited to resolving fine-grained structural and texture ambiguities. 

Second, we revisit classifier-free guidance (CFG) \cite{ho2022classifier} for vision-only restoration problems. In modern diffusion models, CFG is a critical mechanism for boosting perceptual quality and generative sharpness, yet existing vision-only SR methods \cite{sr3,li2022srdiff,yue2023resshift} mainly focus on conditional restoration itself and rarely study guidance as an explicit tool for improving restoration quality. At the same time, directly adopting the standard fully unconditional auxiliary branch is suboptimal for SR: once the LR condition is entirely removed, the auxiliary branch must learn generic generation from scratch, while the conditional branch is left to carry input controllability. This role split is difficult to optimize in restoration, and a poorly learned unconditional branch can undermine generation quality while providing a weaker reference for restoration-oriented guidance. We therefore propose a \textit{restoration-oriented guidance} that replaces the unconditional branch in CFG with a partially conditioned branch. Specifically, weak LR structural cues are preserved while the high-level semantics are dropped. This keeps the auxiliary branch generative yet explicitly input-anchored, making the resulting guidance direction restoration-oriented. Consequently, VOSR strengthens input-consistent restoration rather than amplifying generic generative preference. Interestingly, it also induces a different inference behavior from conventional T2I-based guidance: larger guidance scales favor fidelity by moving predictions closer to the fully LR-conditioned branch, while smaller scales allow stronger generation by leaning toward the partially conditioned branch.

Beyond restoration quality, inference efficiency is also critical for practical SR. We therefore first train a multi-step VOSR model and then distill it into a one-step variant for fast deployment. Fig.~\ref{fig:overview} compares VOSR with generative SR methods in terms of performance, efficiency, and training cost across both T2I-based and vision-only approaches under multi-step and one-step settings. As shown in the figure, VOSR is competitive with or better than most T2I-based SR methods in both regimes, while outperforming significantly previous vision-only methods in perceptual quality. Our multi-step model is substantially more efficient than existing T2I-based SR systems, and the one-step variant remains comparable to recent one-step T2I methods while enabling fast deployment. Moreover, VOSR requires only about one-tenth of the training cost of representative T2I-based SR methods. These results suggest that a restoration-oriented, vision-only framework can simultaneously offer strong perceptual quality, practical efficiency, and much lower training cost without multimodal pretraining.


\section{Related Work}
\label{sec:related}

\textbf{Vision-Only Image Super-Resolution}.
Image super-resolution has traditionally been formulated as a purely visual restoration problem. Early deep models, including convolutional networks~\cite{dong2014learning, rcan, rdn} and vision transformers~\cite{liang2021swinir,zhang2022efficient,chen2023activating}, are usually trained with pixel-wise objectives such as $\ell_1$ loss, which achieve strong distortion-based measures (\eg, PSNR) but often produce over-smoothed textures. To improve perceptual quality, later methods incorporate perceptual losses~\cite{lpips} or adversarial training~\cite{wang2018esrgan,zhang2021designing,wang2021real}, but these approaches often suffer from training instability and visible artifacts.
Diffusion models~\cite{ho2020denoising,song2020score} have recently emerged as a powerful alternative for generative SR. Methods such as SR3~\cite{sr3}, SRDiff~\cite{li2022srdiff}, ResShift~\cite{yue2023resshift}, and SinSR~\cite{wang2023sinsr} show that high-quality detail synthesis is possible without multimodal supervision. From a task perspective, this paradigm is naturally aligned with restoration because generation is driven directly by the degraded input. However, existing vision-only generative SR methods mainly condition restoration through structural cues from the LR input, without explicitly introducing semantic guidance. Such conditioning often provides insufficient high-level information under severe degradation. In other words, the limitation of vision-only SR does not lie in its task formulation but its lack of sufficiently strong semantic abstraction and restoration-oriented guidance.

\noindent\textbf{T2I-based Image Super-Resolution}.
The success of large-scale T2I diffusion models and related generative foundation models in a broad range of visual generation tasks~\cite{rombach2022high, podell2023sdxl, esser2024scaling, zhang2023adding, tao2025instantcharacter, yang2026effectmaker} has inspired a new line of SR methods that adapt pretrained T2I backbones for restoration~\cite{wang2023exploiting,wu2024seesr,sun2023improving,yang2023pixel,sun2024pixel,qu2024xpsr,ai2024dreamclear,chen2025faithdiff,duan2025dit4sr}. These methods benefit from strong priors learned from massive image-text corpora and often achieve impressive perceptual quality. Typical designs inject LR information through prompts, ControlNet-like branches, adapters, or text-aligned features.

Despite their success, T2I-based SR adopts a rather different training paradigm from native image restoration. Rather than directly training a restoration model, these methods constrain a generic image generator to comply with the LR image. While highly effective in leveraging pre-trained image priors, they introduce a structural tension between generic image generation and faithful restoration. Moreover, the semantic cues used by these methods are often represented in text or text-aligned spaces, which are spatially coarse to align with pixel-level image details. In addition, this paradigm inherits the high training and deployment cost of multimodal foundation models.

\noindent\textbf{Toward Restoration-Oriented Generative SR.}
The above two paradigms reveal an important issue in current generative SR research. On the one hand, vision-only SR is naturally grounded in the degraded input and better aligned with the restoration objective, yet its generative capability is limited by weak semantic abstraction under severe degradation. On the other hand, T2I-based SR provides powerful image priors, yet its generic multimodal generation framework is not native to restoration and is often constrained by the model size of the pretrained backbone, making lightweight, budget-aware restoration design challenging.

These observations raise an inspiring question: \textit{can we train a generative SR model that remains fully grounded in the input LR image while incorporating stronger semantics directly in the visual domain?} Meanwhile, recent one-step and few-step acceleration methods in generative modeling~\cite{song2023consistency, frans2024one, yin2024one, lu2024simplifying, geng2025mean}, together with emerging efficient SR approaches~\cite{wu2024one, sun2024pixel, chen2024adversarial, dong2025tsd, yi2025fine, yue2024arbitrary}, indicate that generative SR should ideally support both high perceptual quality and efficient inference. In this paper, we develop a vision-only generative SR framework along this line. 

\section{Method}
\label{sec:method}\vspace{-1mm}
\setlength{\abovedisplayskip}{4pt}
\setlength{\belowdisplayskip}{4pt}
\setlength{\abovedisplayshortskip}{2pt}
\setlength{\belowdisplayshortskip}{2pt}

\begin{figure*}[t]
    \centering
    \includegraphics[width=0.93\linewidth]{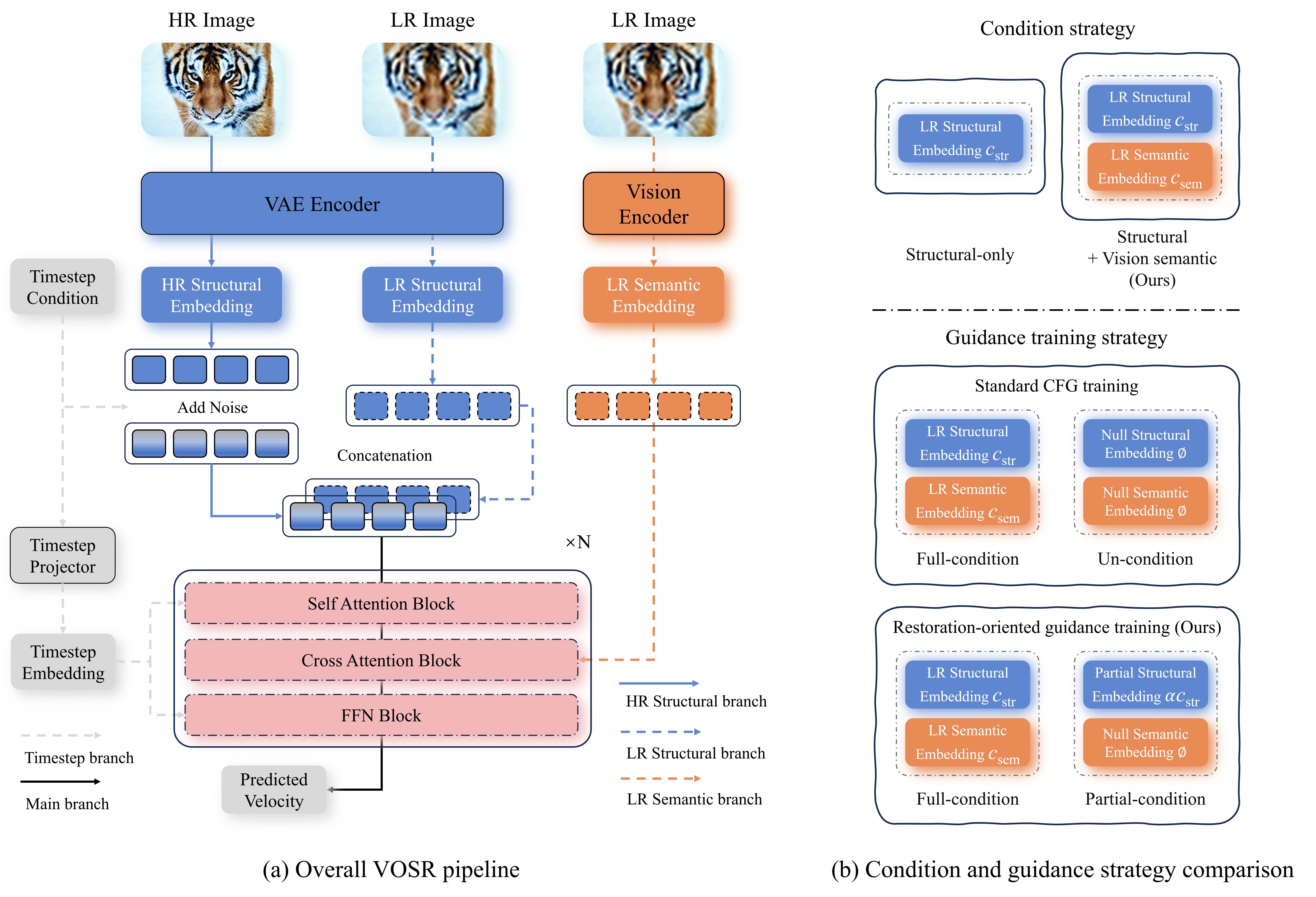}
    \vspace{-3mm}
    \caption{Overview of VOSR. (a) Framework overview. Given an LR image, VOSR builds two complementary conditions from the input: a spatially aligned structural condition in the VAE latent space and a high-level visual semantic condition extracted by a pretrained vision encoder. These conditions are injected into a diffusion transformer to predict the denoising velocity for HR reconstruction. (b) Condition and guidance design. Compared with prior vision-only SR that mainly relies on structural-only conditioning, VOSR introduces an additional visual semantic condition to reduce semantic ambiguity in restoration. Moreover, instead of using a fully unconditional branch as in standard classifier-free guidance, VOSR adopts a restoration-oriented guidance and removes semantic guidance while retaining weakened LR structural cues, making the guidance input-anchored and more suitable for restoration.}
    \label{fig:arch}
    \vspace{-4mm}
\end{figure*}

\subsection{Framework Overview}
Given a low-resolution image $I_{\mathrm{LR}}$, the goal of SR is to recover a high-resolution image $I_{\mathrm{HR}}$ that is perceptually realistic and faithful to the input. Unlike text-to-image generation, SR is purely vision-conditioned. Based on this property, we build \textbf{VOSR}, a vision-only generative SR framework, whose overall pipeline is shown in Fig.~\ref{fig:arch}(a). 

Following latent diffusion~\cite{rombach2022high}, we encode the target HR image into a latent code $z_0=\mathcal{E}(I_{\mathrm{HR}})$ with a VAE encoder $\mathcal{E}$, sample Gaussian noise $z_1\sim\mathcal{N}(0,\mathbf{I})$, and define the intermediate latent at time $t\in[0,1]$ as $z_t=(1-t)z_0+t z_1$. The diffusion backbone then predicts the velocity target $v_t=z_1-z_0$ from $z_t$ under two LR-derived conditions: \textit{low-level structural conditioning} and \textit{high-level visual semantic guidance}. Specifically, we encode the LR image into the same latent space as a spatially aligned structural condition to preserve content and layout, and extract semantic features from a pretrained vision encoder as visual semantic guidance to provide semantically rich and spatially grounded cues for detail generation.  

Fig.~\ref{fig:arch}(b) highlights the proposed condition and guidance strategies. For the condition design, prior vision-only SR mainly relies on structural-only conditioning \cite{sr3, yue2023resshift}. In contrast, VOSR further introduces visual semantic embedding. The two conditions are complementary: the structural condition preserves fidelity, while semantic guidance helps resolve ambiguity in HR reconstruction. For the guidance strategy during training, instead of removing all conditions in the unconditional branch in standard classifier-free guidance, we propose a partially conditioned branch, where we remove semantic guidance and weaken the LR structural guidance. This keeps the guidance anchored to the input and better suits restoration.

Built on the above design, we first train a multi-step model for high-quality generation and then distill it into a one-step model for efficient inference. In this way, VOSR unifies vision semantic condition, restoration-oriented guidance, and one-step distillation in a single vision-only SR framework, as introduced in detail below.

\subsection{Vision Semantic Condition}
\label{sec:sem}
We first specify the two conditioning strategies in Fig.~\ref{fig:arch}(b). Existing vision-only generative SR models \cite{sr3, yue2023resshift} mainly rely on structural cues from the LR input. However, under severe degradation, structural conditioning alone can be insufficient because the LR observation admits multiple plausible HR reconstructions, causing semantic ambiguity. VOSR therefore adopts a two-branch conditioning design that combines LR structure with semantic information.

Specifically, the structural branch encodes the LR image with the same VAE encoder $\mathcal{E}$ as used to map HR images into the latent space, producing a structural condition $c_{\mathrm{str}}=\mathcal{E}(I_{\mathrm{LR}})$ to preserve spatial content. In parallel, a pretrained vision encoder $\mathcal{V}$, such as DINO \cite{oquab2023dinov2, simeoni2025dinov3}, extracts a semantic condition $c_{\mathrm{sem}}=\mathcal{V}(I_{\mathrm{LR}})$. Unlike text or text-aligned representations, $c_{\mathrm{sem}}$ is extracted entirely in the visual domain and is better suited to fine-grained restoration. Our denoising backbone is a diffusion transformer \cite{yao2025reconstruction} that takes the noisy HR latent $z_t$ and timestep embedding as input. The structural condition $c_{\mathrm{str}}$ is injected as spatially aligned latent conditioning, while the semantic condition $c_{\mathrm{sem}}$ is introduced through cross-attention to provide high-level context. In this way, structural features preserve spatial fidelity, while semantic features resolve ambiguity during detail synthesis. This design introduces high-level semantics entirely within the visual domain, avoiding text or text-aligned conditioning in SR.

\subsection{Restoration-Oriented Guidance}
\label{sec:partial}
After defining the visual conditions, we discuss how to guide generation toward perceptually realistic yet input-faithful restoration. Classifier-free guidance (CFG)~\cite{ho2022classifier} is a key technique in modern conditional diffusion models, as it can substantially improve sample sharpness and controllability at inference time. However, despite its importance in conditional generation, CFG remains largely underexplored in existing vision-only SR methods~\cite{sr3, li2022srdiff, yue2023resshift}. We argue that, for restoration, the design of the auxiliary guidance branch is especially important, because SR is not text-conditioned generation but input-anchored generation.

We examine the standard CFG for SR, as shown in Fig.~\ref{fig:arch}(b). 
During training, the model randomly switches between a fully conditioned branch and an unconditional branch. At inference, the guided prediction is computed as
$v_{\mathrm{cfg}}=v_{\mathrm{uncond}}+s\left(v_{\mathrm{cond}}-v_{\mathrm{uncond}}\right)$,
where $v_{\mathrm{cond}}$ and $v_{\mathrm{uncond}}$ denote the conditional and unconditional predictions, respectively, and $s$ is the guidance scale. This formulation is highly effective in T2I generation, where the unconditional branch provides a natural reference for generic image generation without text control. For SR trained from scratch, however, such a fully unconditional branch is less suitable. Once the LR input is entirely removed, the auxiliary branch must learn generic image generation, while the conditional branch is solely responsible for maintaining fidelity to the input. This separation of roles makes it difficult to optimize for image restoration, and weak unconditional learning may provide a poor guidance reference.

Based on this observation, we replace the unconditional branch with a \emph{partially conditioned} branch. The goal is not to drop conditions arbitrarily, but to construct an auxiliary branch, which is different from the fully conditioned branch and defines a meaningful restoration-oriented guidance direction. Specifically, instead of removing all conditions, we retain weakened LR structural cues while dropping semantic guidance. Compared with the fully conditioned branch using both structural and semantic conditions $(c_{\mathrm{str}}, c_{\mathrm{sem}})$, the partially conditioned branch keeps only a scaled structural condition and removes semantic conditioning. Let $\alpha\in(0,1)$ denote the structural retention factor. The two branches are defined as:
$v_{\mathrm{cond}}=v_\theta(z_t,t,c_{\mathrm{str}},c_{\mathrm{sem}})$
and
$v_{\mathrm{pcond}}=v_\theta(z_t,t,\alpha c_{\mathrm{str}},\varnothing)$,
where $\varnothing$ denotes the absence of semantic conditioning. In this way, both branches remain anchored to the LR input, while their difference captures the effect of stronger structure and semantic guidance. During training, we randomly sample either the fully conditioned mode or the partially conditioned mode so that the shared model learns both behaviors under a unified objective. Following standard latent diffusion training in velocity parameterization, the multi-step objective is
\begin{equation}
\mathcal{L}_{\mathrm{ms}}=\mathbb{E}_{z_0,z_1,t,\kappa}\left[\left\|v_\theta(z_t,t,\kappa)-v_t\right\|_2^2\right],
\end{equation}
where $v_t=z_1-z_0$ is the ground-truth velocity target, and $\kappa$ denotes the sampled conditioning mode, \ie, either the fully conditioned branch $(c_{\mathrm{str}}, c_{\mathrm{sem}})$ or the partially conditioned branch $(\alpha c_{\mathrm{str}}, \varnothing)$. 

At test time, we apply restoration-oriented guidance:
\begin{equation}
v_{\mathrm{cfg}}=v_{\mathrm{pcond}}+s\left(v_{\mathrm{cond}}-v_{\mathrm{pcond}}\right),
\end{equation}
where $s$ is the guidance scale. Unlike standard CFG, the auxiliary branch here is still weakly conditioned on the LR input. Therefore, the guidance direction does not move from unconditional generation toward conditioned restoration; instead, it moves from a weakly anchored branch toward a strongly anchored one. As $s$ increases, the prediction is driven closer to the fully conditioned branch, which strengthens input consistency. As $s$ decreases, the prediction leans toward the partially conditioned branch, where semantic conditioning is removed and structural conditioning is weakened, thereby allowing a larger space of plausible detail generation. This behavior is notably different from the CFG commonly used in T2I-based SR~\cite{wu2024seesr, chen2025faithdiff}, which can be written as:
\begin{equation}
\begin{aligned}
v_{\mathrm{cfg}} ={}& v_\theta(z_t,t,c_{\mathrm{lr}},\varnothing) \\
&+ s\Big( v_\theta(z_t,t,c_{\mathrm{lr}},c_{\mathrm{sem}})
- v_\theta(z_t,t,c_{\mathrm{lr}},\varnothing) \Big),
\end{aligned}
\end{equation}
where the LR control is shared by both branches. The  direction of guidance mainly reflects the injection of additional semantic priors, and increasing the guidance scale usually amplifies generative semantics. In our formulation, by contrast, $s=0$ recovers the partially conditioned branch and $s=1$ the fully conditioned branch. Hence, a smaller $s$ tends to produce more generative results, whereas a larger $s$ favors better fidelity to the LR input. This makes our VOSR restoration-oriented, which balances generation and fidelity within an input-anchored restoration space.

\subsection{One-Step Distillation for Efficient Inference}
Although the multi-step VOSR model provides strong restoration quality, iterative sampling is still expensive for practical deployment. We therefore distill it into a fast student while preserving the same vision semantic condition and restoration-oriented guidance as the teacher. That is, distillation changes only sampling efficiency, not the restoration formulation.

We parameterize the student as $f_{\theta}(z_t, t, r, \kappa)$, where the auxiliary target time $r$ denotes the target timestep to be predicted by the student, unifying local prediction ($r=t$) and compressed cross-time prediction ($r<t$). To keep the conditioning interface consistent with Sec.~\ref{sec:partial}, the student is trained under the same conditioning mode as the teacher, \ie, either the fully conditioned branch $(c_{\mathrm{str}}, c_{\mathrm{sem}})$ or the partially conditioned branch $(\alpha c_{\mathrm{str}}, \varnothing)$, denoted collectively by $\kappa$. The teacher target also remains restoration-oriented: we use the guided prediction $v_{\mathrm{tea}}^{\mathrm{guide}} = v_{\mathrm{pcond}} + \omega \left( v_{\mathrm{cond}} - v_{\mathrm{pcond}} \right)$, where $v_{\mathrm{cond}}$ and $v_{\mathrm{pcond}}$ are the fully conditioned and partially conditioned teacher predictions, respectively, and $\omega$ is the distillation-time guidance weight, distinct from the inference-time guidance scale $s$ in Sec.~\ref{sec:partial}. The shared base objective is:
\begin{equation}
\mathcal{L}_{\mathrm{base}} =
\mathbb{E}_{z_0,\epsilon,t,\kappa}\left[
\left\| f_{\theta}(z_t, t, r{=}t, \kappa) - v_{\mathrm{tea}}^{\mathrm{guide}} \right\|^2
\right],
\end{equation}
which transfers the same input-anchored generative restoration capability from teacher to student.

Existing one-step distillation methods differ substantially in memory cost. To support model scaling, we focus on memory-friendly variants and exclude designs that require multiple coupled networks or Jacobian-vector product (JVP) evaluations, which substantially increase memory cost in our setting. Under this criterion, we study two task-adapted variants inspired by shortcut-based and recursive-consistency-based acceleration methods~\cite{frans2024one,sun2026anystep}. Both variants share the same conditioning interface and base objective, and differ only in the auxiliary distillation loss used to compress the denoising trajectory. Empirically, the recursive-consistency-based variant performs best for SR, offering a better balance between perceptual quality and structural fidelity. We therefore adopt this variant in our main experiments. In this way, VOSR preserves the same restoration-oriented formulation as the multi-step teacher while enabling efficient one-step inference. The detailed distillation formulations are provided in the \textbf{Appendix}.

\vspace{-1mm}
\section{Experiments}
\label{sec:exp}

\subsection{Experimental Settings}

\noindent
\textbf{Training and Testing Datasets}. 
To construct a diverse training corpus, we collect web images and apply quality and diversity filtering, including gradient-based filtering, image-entropy filtering, resolution filtering, and category-level de-duplication and balancing, obtaining about 100M images. We then synthesize LR-HR training pairs using the Real-ESRGAN degradation pipeline~\cite{wang2021real}. VOSR is trained solely on the synthetic pairs, but is evaluated on both synthetic and real-world test sets to assess generalization.

For the synthetic test set, we use the validation split of LSDIR~\cite{li2023lsdir}, which contains 250 images. We synthesize LR inputs using the same Real-ESRGAN degradation pipeline as in training.
Real-world paired test sets are more valuable for SR because they enable zero-shot evaluation under unknown degradations.
For real-world evaluation, we first use the well-known RealSR~\cite{realsr} benchmark. However, the quality of GT images in this benchmark is relatively out-of-date. Therefore, we construct a new test set, namely \textbf{ScreenSR}, through a screen re-photography pipeline. ScreenSR provides higher-quality references and more diverse content than existing real-world paired benchmarks, covering a broad range of scenes and scales. Detailed data construction process, capture settings, devices, and GT quality comparisons are provided in the \textbf{Appendix}.

For LSDIR and RealSR, we extract $512\times512$ center-cropped GT patches and generate the $128\times128$ LR inputs, forming a standardized $\times4$ SR protocol. For ScreenSR, both the LQ input and GT image are of size $512\times512$. 


\vspace{+1mm}
\noindent
\textbf{Compared Methods}. 
We compare VOSR with representative generative SR methods under multi-step and one-step settings, including vision-only (VO) and T2I-based approaches. The multi-step baselines include the VO method ResShift~\cite{yue2023resshift} and the T2I-based methods StableSR~\cite{wang2023exploiting}, PASD~\cite{yang2023pixel}, SeeSR~\cite{wu2024seesr}, and DiT4SR~\cite{duan2025dit4sr}. For one-step comparison, we include the VO method SinSR~\cite{wang2023sinsr} and the T2I-based methods OSEDiff~\cite{wu2024one} and PiSA-SR~\cite{sun2024pixel}. These baselines cover the main generative SR paradigms, enabling comparison of restoration quality and efficiency between VOSR and both VO and T2I-based methods.

\vspace{+1mm}
\noindent
\textbf{Evaluation Metrics}.
We evaluate all methods with full-reference and no-reference metrics. For distortion fidelity, we report PSNR and SSIM~\cite{ssim} on the Y channel of the YCbCr color space. For reference-based perceptual quality, we use LPIPS~\cite{lpips}, DISTS~\cite{dists}, and AFINE-FR~\cite{chen2025toward}. For no-reference perceptual quality, we report NIQE~\cite{niqe}, MUSIQ~\cite{musiq}, MANIQA~\cite{maniqa}, AFINE-NR~\cite{chen2025toward}, and TOPIQ-NR~\cite{chen2024topiq}. We also report a user study in the \textbf{Appendix}.

\vspace{+1mm}
\noindent
\textbf{Implementation Details}.
We train two VOSR variants of different sizes, both using LightningDiT \cite{yao2025reconstruction} as the backbone. Their diffusion models contain 0.5B and 1.4B parameters, denoted as VOSR-0.5B and VOSR-1.4B, respectively. For each size, we report both a multi-step model and its distilled one-step counterpart, denoted by the suffixes ``-ms'' and ``-os'' (\eg, VOSR-0.5B-ms and VOSR-0.5B-os). VOSR-0.5B uses the SD2.1 VAE and, following AdcSR \cite{chen2024adversarial}, a retrained lightweight decoder for SR to reduce peak inference memory with minimal impact on decoding quality. VOSR-1.4B instead adopts a 16-channel latent VAE \cite{wu2025qwenimagetechnicalreport} to reduce the irreversible information loss of the standard 4-channel compression. For semantic encoding, VOSR-0.5B uses DINOv2-Base and VOSR-1.4B uses DINOv2-Large~\cite{oquab2023dinov2}. At inference, the multi-step model uses 25 sampling steps. Detailed training hyperparameters and architecture configurations are provided in the \textbf{Appendix}.

\begin{table*}[t]
\centering
\caption{Quantitative results on LSDIR, ScreenSR, and RealSR. Methods are grouped into multi-step (ms) and one-step (os) settings. T2I and VO denote text-to-image-based and vision-only methods, respectively. $\uparrow$ ($\downarrow$) indicates higher (lower) is better. T2I-based methods are marked in \textcolor{gray}{gray} for visual distinction. The best and second-best results are highlighted in \textbf{\textcolor{red}{bold red}} and \textbf{\textcolor{blue}{bold blue}}, respectively.}
\vspace{-2mm}
\resizebox{\linewidth}{!}{
\begin{tabular}{@{}cccccccccccccc@{}}
\toprule
Dataset & Setting & Method & Type & PSNR$\uparrow$ & SSIM$\uparrow$ & LPIPS$\downarrow$ & DISTS$\downarrow$ & AFINE-FR$\downarrow$ & NIQE$\downarrow$ & MUSIQ$\uparrow$ & MANIQA$\uparrow$ & AFINE-NR$\downarrow$ & TOPIQ-NR$\uparrow$ \\
\midrule
\multirow{12}{*}{LSDIR}
& \multirow{7}{*}{Multi-step}
& \textcolor{gray}{StableSR} & \multirow{4}{*}{\textcolor{gray}{T2I}} & \textcolor{gray}{19.1120} & \textcolor{gray}{0.3751} & \textcolor{gray}{0.5169} & \textcolor{gray}{0.3068} & \textcolor{gray}{0.5477} & \textcolor{gray}{5.4703} & \textcolor{gray}{56.2263} & \textcolor{gray}{0.5345} & \textcolor{gray}{-0.7529} & \textcolor{gray}{0.5627} \\
&
& \textcolor{gray}{PASD} &  & \textcolor{gray}{19.0817} & \textcolor{gray}{0.4186} & \textcolor{gray}{0.4898} & \textcolor{gray}{0.2433} & \textcolor{gray}{-0.1689} & \textcolor{gray}{4.4446} & \textcolor{gray}{65.6496} & \textcolor{gray}{0.5780} & \textcolor{gray}{-0.9574} & \textcolor{gray}{0.5827} \\
&
& \textcolor{gray}{SeeSR} &  & \textcolor{gray}{19.1408} & \textcolor{gray}{0.4198} & 0.4180 & 0.2147 & -0.7413 & 4.2406 & 72.3981 & 0.6616 & -1.0766 & \textbf{\textcolor{red}{0.7316}} \\
&
& \textcolor{gray}{DiT4SR} &  & \textcolor{gray}{17.7220} & \textcolor{gray}{0.3833} & \textcolor{gray}{0.4458} & \textcolor{gray}{0.2246} & \textcolor{gray}{-0.7300} & \textbf{\textcolor{blue}{4.2398}} & \textbf{\textcolor{blue}{73.3560}} & \textbf{\textcolor{red}{0.6854}} & \textbf{\textcolor{red}{-1.1859}} & 0.6951 \\
&
& ResShift & \multirow{3}{*}{VO} & \textbf{\textcolor{red}{19.8723}} & 0.4300 & 0.4784 & 0.2699 & 1.0205 & 6.0833 & 59.1250 & 0.5243 & -0.7036 & 0.4966 \\
&
& VOSR-0.5B-ms &  & \textbf{\textcolor{blue}{19.5677}} & \textbf{\textcolor{blue}{0.4312}} & \textbf{\textcolor{blue}{0.3984}} & \textbf{\textcolor{blue}{0.2126}} & \textbf{\textcolor{blue}{-0.8310}} & \textbf{\textcolor{red}{4.0466}} & 73.1033 & 0.6679 & -1.0844 & \textbf{\textcolor{blue}{0.7231}} \\
&
& VOSR-1.4B-ms &  & 19.3896 & \textbf{\textcolor{red}{0.4329}} & \textbf{\textcolor{red}{0.3857}} & \textbf{\textcolor{red}{0.1977}} & \textbf{\textcolor{red}{-0.9525}} & 4.4554 & \textbf{\textcolor{red}{74.0258}} & \textbf{\textcolor{blue}{0.6828}} & \textbf{\textcolor{blue}{-1.0859}} & 0.6631 \\
\cmidrule(lr){2-14}
& \multirow{5}{*}{One-step}
& \textcolor{gray}{OSEDiff} & \multirow{2}{*}{\textcolor{gray}{T2I}} & \textbf{\textcolor{red}{19.7116}} & \textbf{\textcolor{blue}{0.4405}} & \textcolor{gray}{0.3898} & \textcolor{gray}{0.2339} & \textcolor{gray}{-0.3279} & 3.9766 & \textcolor{gray}{69.6531} & \textcolor{gray}{0.6105} & -0.9409 & \textcolor{gray}{0.6394} \\
&
& \textcolor{gray}{PiSA-SR} &  & \textbf{\textcolor{blue}{19.6890}} & \textbf{\textcolor{red}{0.4413}} & \textbf{\textcolor{red}{0.3758}} & 0.2222 & \textcolor{gray}{-0.3143} & \textbf{\textcolor{blue}{3.9612}} & 70.8861 & 0.6349 & \textcolor{gray}{-0.9395} & \textbf{\textcolor{blue}{0.6699}} \\
&
& SinSR & \multirow{3}{*}{VO} & 19.6740 & 0.4174 & 0.4562 & 0.2540 & 1.3025 & 5.3233 & 61.8681 & 0.5076 & -0.6218 & 0.5435 \\
&
& VOSR-0.5B-os &  & 19.3893 & 0.4300 & 0.3888 & \textbf{\textcolor{blue}{0.2079}} & \textbf{\textcolor{red}{-0.7175}} & \textbf{\textcolor{red}{3.6476}} & \textbf{\textcolor{blue}{72.4218}} & \textbf{\textcolor{blue}{0.6529}} & \textbf{\textcolor{red}{-1.0292}} & \textbf{\textcolor{red}{0.6985}} \\
&
& VOSR-1.4B-os &  & 19.2999 & 0.4311 & \textbf{\textcolor{blue}{0.3802}} & \textbf{\textcolor{red}{0.1994}} & \textbf{\textcolor{blue}{-0.5831}} & 4.1093 & \textbf{\textcolor{red}{73.7889}} & \textbf{\textcolor{red}{0.6570}} & \textbf{\textcolor{blue}{-0.9919}} & 0.6525 \\
\midrule
\multirow{12}{*}{ScreenSR}
& \multirow{7}{*}{Multi-step}
& \textcolor{gray}{StableSR} & \multirow{4}{*}{\textcolor{gray}{T2I}} & \textcolor{gray}{21.2073} & \textcolor{gray}{0.6157} & \textcolor{gray}{0.2357} & \textcolor{gray}{0.1618} & \textcolor{gray}{-1.5049} & \textcolor{gray}{4.3150} & \textcolor{gray}{71.8000} & \textcolor{gray}{0.7002} & \textcolor{gray}{-1.1659} & \textcolor{gray}{0.6712} \\
&
& \textcolor{gray}{PASD} &  & 22.2532 & \textcolor{gray}{0.6115} & \textcolor{gray}{0.2526} & 0.1547 & \textcolor{gray}{-1.1288} & \textbf{\textcolor{red}{3.9197}} & \textcolor{gray}{70.2892} & \textcolor{gray}{0.6611} & \textcolor{gray}{-1.0313} & \textcolor{gray}{0.6569} \\
&
& \textcolor{gray}{SeeSR} &  & \textbf{\textcolor{blue}{22.4501}} & \textbf{\textcolor{blue}{0.6275}} & \textbf{\textcolor{blue}{0.2233}} & \textbf{\textcolor{red}{0.1371}} & \textbf{\textcolor{red}{-1.5961}} & 4.1958 & \textcolor{gray}{72.3075} & \textcolor{gray}{0.6900} & \textcolor{gray}{-1.1763} & \textbf{\textcolor{red}{0.7362}} \\
&
& \textcolor{gray}{DiT4SR} &  & \textcolor{gray}{20.8880} & \textcolor{gray}{0.6008} & \textcolor{gray}{0.2513} & \textcolor{gray}{0.1577} & \textcolor{gray}{-1.5020} & \textbf{\textcolor{blue}{4.1910}} & \textcolor{gray}{71.3778} & 0.7091 & \textcolor{gray}{-1.1719} & \textcolor{gray}{0.6761} \\
&
& ResShift & \multirow{3}{*}{VO} & \textbf{\textcolor{red}{23.1442}} & \textbf{\textcolor{red}{0.6622}} & \textbf{\textcolor{red}{0.2198}} & 0.1531 & -0.9480 & 5.1905 & 68.3242 & 0.6250 & -0.9865 & 0.6483 \\
&
& VOSR-0.5B-ms &  & 21.6726 & 0.5959 & 0.2484 & 0.1543 & -1.5410 & 4.3134 & \textbf{\textcolor{blue}{72.7227}} & \textbf{\textcolor{blue}{0.7111}} & \textbf{\textcolor{red}{-1.2834}} & \textbf{\textcolor{blue}{0.7055}} \\
&
& VOSR-1.4B-ms &  & 21.7910 & 0.6013 & 0.2526 & \textbf{\textcolor{blue}{0.1513}} & \textbf{\textcolor{blue}{-1.5591}} & 4.7540 & \textbf{\textcolor{red}{73.3143}} & \textbf{\textcolor{red}{0.7197}} & \textbf{\textcolor{blue}{-1.2577}} & 0.6825 \\
\cmidrule(lr){2-14}
& \multirow{5}{*}{One-step}
& \textcolor{gray}{OSEDiff} & \multirow{2}{*}{\textcolor{gray}{T2I}} & \textcolor{gray}{22.1149} & \textcolor{gray}{0.6303} & \textcolor{gray}{0.2302} & 0.1527 & -1.5546 & \textbf{\textcolor{blue}{4.0806}} & \textcolor{gray}{71.4260} & \textcolor{gray}{0.6809} & \textbf{\textcolor{blue}{-1.2141}} & \textcolor{gray}{0.6281} \\
&
& \textcolor{gray}{PiSA-SR} &  & \textbf{\textcolor{blue}{22.2142}} & \textbf{\textcolor{blue}{0.6415}} & \textbf{\textcolor{red}{0.1951}} & \textbf{\textcolor{red}{0.1384}} & \textbf{\textcolor{blue}{-1.6439}} & \textcolor{gray}{4.2179} & \textbf{\textcolor{red}{72.8114}} & \textbf{\textcolor{red}{0.7265}} & \textbf{\textcolor{red}{-1.2751}} & \textbf{\textcolor{blue}{0.6810}} \\
&
& SinSR & \multirow{3}{*}{VO} & \textbf{\textcolor{red}{23.2057}} & \textbf{\textcolor{red}{0.6542}} & 0.2260 & 0.1590 & -0.2707 & 4.7168 & 67.3158 & 0.5896 & -0.8861 & 0.6156 \\
&
& VOSR-0.5B-os &  & 21.9381 & 0.6142 & \textbf{\textcolor{blue}{0.2198}} & \textbf{\textcolor{blue}{0.1405}} & \textbf{\textcolor{red}{-1.6625}} & \textbf{\textcolor{red}{4.0385}} & 71.9349 & 0.7043 & -1.2042 & \textbf{\textcolor{red}{0.6949}} \\
&
& VOSR-1.4B-os &  & 21.8084 & 0.6182 & 0.2199 & 0.1406 & -1.5984 & 4.5163 & \textbf{\textcolor{blue}{72.7631}} & \textbf{\textcolor{blue}{0.7060}} & -1.1429 & 0.6622 \\
\midrule
\multirow{12}{*}{RealSR}
& \multirow{7}{*}{Multi-step}
& \textcolor{gray}{StableSR} & \multirow{4}{*}{\textcolor{gray}{T2I}} & \textcolor{gray}{24.6426} & \textcolor{gray}{0.7079} & \textcolor{gray}{0.3004} & \textbf{\textcolor{blue}{0.2140}} & \textbf{\textcolor{blue}{-0.7706}} & \textcolor{gray}{5.8838} & \textcolor{gray}{65.8802} & \textcolor{gray}{0.6229} & \textcolor{gray}{-1.0117} & \textcolor{gray}{0.5747} \\
&
& \textcolor{gray}{PASD} &  & 25.2423 & \textbf{\textcolor{blue}{0.7223}} & \textbf{\textcolor{blue}{0.2988}} & \textbf{\textcolor{red}{0.2065}} & \textbf{\textcolor{red}{-0.8323}} & \textbf{\textcolor{red}{5.2047}} & \textcolor{gray}{65.3484} & \textcolor{gray}{0.5960} & \textcolor{gray}{-0.9395} & \textcolor{gray}{0.5811} \\
&
& \textcolor{gray}{SeeSR} &  & \textcolor{gray}{25.1480} & \textcolor{gray}{0.7211} & \textcolor{gray}{0.3007} & 0.2224 & \textcolor{gray}{-0.7431} & \textbf{\textcolor{blue}{5.3973}} & \textbf{\textcolor{blue}{69.8179}} & 0.6451 & \textcolor{gray}{-1.0370} & \textbf{\textcolor{red}{0.6891}} \\
&
& \textcolor{gray}{DiT4SR} &  & \textcolor{gray}{23.5956} & \textcolor{gray}{0.6722} & \textcolor{gray}{0.3096} & 0.2224 & \textcolor{gray}{-0.5738} & \textcolor{gray}{6.2377} & \textcolor{gray}{67.3127} & \textbf{\textcolor{red}{0.6564}} & \textbf{\textcolor{red}{-1.0849}} & 0.5982 \\
&
& ResShift & \multirow{3}{*}{VO} & \textbf{\textcolor{red}{26.2630}} & \textbf{\textcolor{red}{0.7405}} & 0.3468 & 0.2495 & -0.6103 & 7.1790 & 58.4687 & 0.5343 & -0.8146 & 0.4891 \\
&
& VOSR-0.5B-ms &  & \textbf{\textcolor{blue}{25.4361}} & 0.7125 & 0.3069 & 0.2260 & -0.7413 & 5.7070 & 68.9277 & 0.6429 & -1.0620 & \textbf{\textcolor{blue}{0.6554}} \\
&
& VOSR-1.4B-ms &  & 25.2886 & 0.7150 & \textbf{\textcolor{red}{0.2961}} & 0.2226 & -0.7618 & 6.2614 & \textbf{\textcolor{red}{70.4718}} & \textbf{\textcolor{blue}{0.6510}} & \textbf{\textcolor{blue}{-1.0847}} & 0.6354 \\
\cmidrule(lr){2-14}
& \multirow{5}{*}{One-step}
& \textcolor{gray}{OSEDiff} & \multirow{2}{*}{\textcolor{gray}{T2I}} & \textcolor{gray}{25.1517} & 0.7341 & \textcolor{gray}{0.2920} & \textcolor{gray}{0.2128} & -0.7169 & \textcolor{gray}{5.6401} & 69.0830 & 0.6335 & \textbf{\textcolor{blue}{-1.0489}} & 0.6253 \\
&
& \textcolor{gray}{PiSA-SR} &  & \textbf{\textcolor{blue}{25.5030}} & \textbf{\textcolor{red}{0.7418}} & \textbf{\textcolor{red}{0.2672}} & \textbf{\textcolor{red}{0.2044}} & -0.7713 & 5.5033 & \textbf{\textcolor{blue}{70.1421}} & \textbf{\textcolor{red}{0.6551}} & \textbf{\textcolor{red}{-1.0699}} & 0.6373 \\
&
& SinSR & \multirow{3}{*}{VO} & \textbf{\textcolor{red}{26.2766}} & \textbf{\textcolor{blue}{0.7347}} & 0.3188 & 0.2352 & -0.4477 & 6.2900 & 60.7849 & 0.5413 & -0.7489 & 0.5160 \\
&
& VOSR-0.5B-os &  & 25.4189 & 0.7220 & 0.2856 & 0.2110 & \textbf{\textcolor{blue}{-0.9520}} & \textbf{\textcolor{red}{5.2790}} & 69.7775 & 0.6347 & -0.9831 & \textbf{\textcolor{red}{0.6719}} \\
&
& VOSR-1.4B-os &  & 25.2284 & 0.7175 & \textbf{\textcolor{blue}{0.2732}} & \textbf{\textcolor{blue}{0.2054}} & \textbf{\textcolor{red}{-0.9951}} & \textbf{\textcolor{blue}{5.4303}} & \textbf{\textcolor{red}{70.5813}} & \textbf{\textcolor{blue}{0.6443}} & -1.0109 & \textbf{\textcolor{blue}{0.6392}} \\
\bottomrule
\end{tabular}
}
\label{tab:tab-main}
\vspace{-3mm}
\end{table*}

\subsection{Experimental Results}
\noindent
\textbf{Quantitative Comparisons}. Table~\ref{tab:tab-main} reports results on LSDIR, ScreenSR, and RealSR under both multi-step (ms) and one-step (os) settings. VOSR consistently outperforms previous vision-only methods and remains competitive with strong T2I-based methods, especially in perceptual metrics. On LSDIR, VOSR achieves the best overall perceptual performance: in the multi-step setting, VOSR-1.4B-ms obtains the best LPIPS, DISTS, AFINE-FR, and MUSIQ, while VOSR-0.5B-ms achieves the best NIQE; in the one-step setting, both VOSR variants are highly competitive, with VOSR-0.5B-os achieving the best NIQE, AFINE-NR, and TOPIQ-NR. On the real-world benchmarks ScreenSR and RealSR, VOSR also performs strongly in perceptual quality without targeting the highest PSNR. For example, VOSR-1.4B-ms achieves the best MUSIQ and MANIQA on ScreenSR and the best LPIPS and MUSIQ on RealSR, while the one-step VOSR models remain competitive on AFINE-FR, AFINE-NR, DISTS, and TOPIQ-NR. Compared with prior vision-only methods such as ResShift and SinSR, VOSR yields much better perceptual quality while remaining competitive in distortion fidelity. While achieving competitive quantitative results with those strong T2I-based competitors, our qualitative comparisons (see the next paragraph) further show that VOSR can better preserve subtle local details, such as small characters and weak textures. These results suggest that without multimodal pretraining, a native vision-only SR model can achieve even better performance with more balanced perceptual quality and faithfulness than T2I-based alternatives.

\begin{figure*}[!t]
    \centering
    \includegraphics[width=\linewidth]{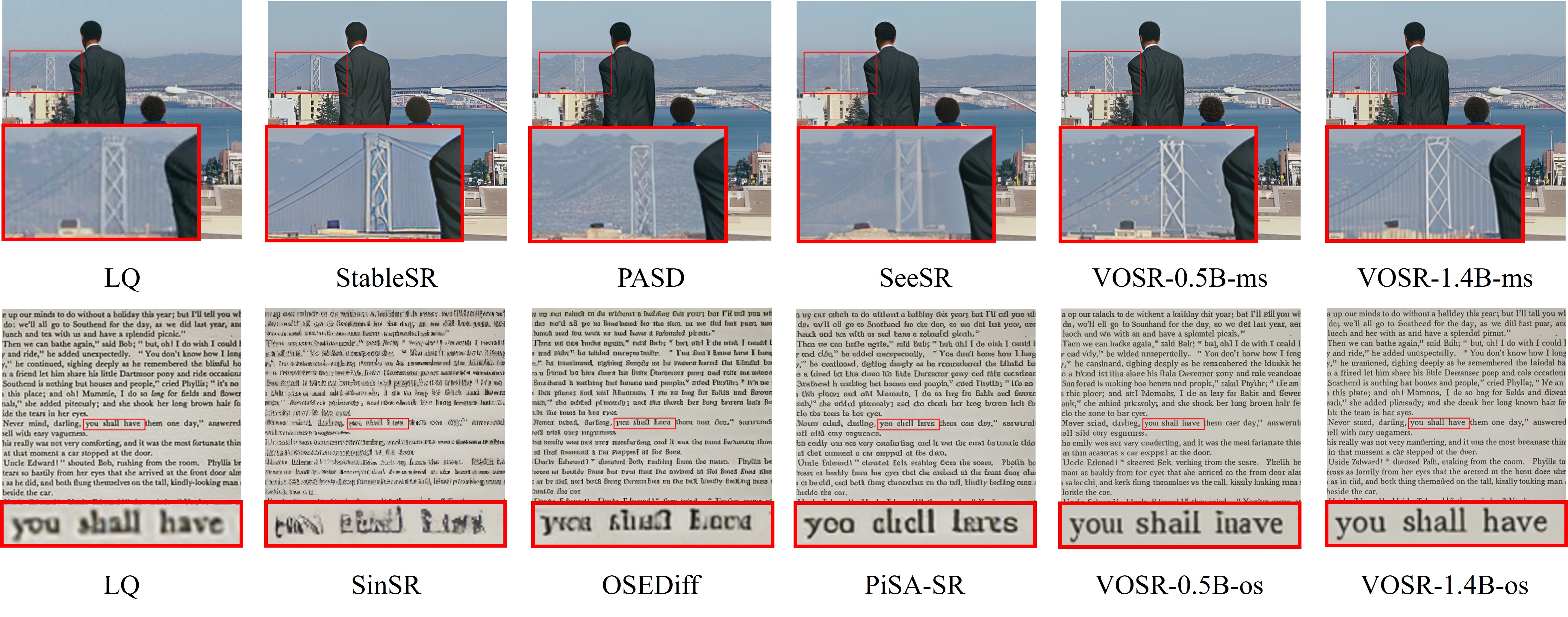}
    \vspace{-7mm}
    \caption{Multi-step (top) and one-step (bottom) SR visual comparison on RealDeg \cite{chen2025faithdiff} \texttt{Cf/0020.png} and ScreenSR \texttt{010.png}.}
    \label{fig:vis-comp}
    \vspace{-4mm}
\end{figure*}

\vspace{-1mm}
\noindent
\textbf{Qualitative Comparisons}. Fig.~\ref{fig:vis-comp} provides visual comparisons under both multi-step and one-step settings. In the first row, the bridge structures are challenging because the tower and cables can be easily oversmoothed. Compared with multi-step T2I-based methods, VOSR restores these fine structures more clearly and faithfully, while competing methods either smooth them out or recover them only partially. In the second row, SinSR shows weak generative capability and recovers character shapes incompletely. One-step T2I-based methods produce sharper letters but introduce inaccurate glyph structures. By contrast, VOSR restores clearer, more faithful details in both cases. We attribute these advantages to the restoration-oriented design of VOSR: instead of using coarse text-aligned semantics from the T2I generative space, VOSR uses denser visual semantics grounded to the LR input, making it more effective at enhancing subtle local details while maintaining structural fidelity. Our results show that a vision-only, restoration-oriented framework can  balance between perceptual quality and faithfulness for SR. More visual comparisons are provided in the \textbf{Appendix}.

\begin{table*}[!t]
\centering
\scriptsize
\caption{Complexity comparison on $512 \times 512$ inputs. Runtime is measured on an A100 GPU with batch size 1 in FP16.}
\label{tab:complexity}
\vspace{-2mm}
\fontsize{7pt}{8.4pt}\selectfont
\setlength{\tabcolsep}{3.2pt}
\renewcommand{\arraystretch}{1.1}
\begin{tabular*}{\textwidth}{@{\extracolsep{\fill}}l
    ccc
    cccc
    ccc
    cc@{}}
\toprule
 & \multicolumn{3}{c}{\textit{Multi-step $\cdot$ Vision-only}}
 & \multicolumn{4}{c}{\textit{Multi-step $\cdot$ T2I-based}}
 & \multicolumn{3}{c}{\textit{One-step $\cdot$ Vision-only}}
 & \multicolumn{2}{c}{\textit{One-step $\cdot$ T2I-based}} \\
\cmidrule(lr){2-4}\cmidrule(lr){5-8}\cmidrule(lr){9-11}\cmidrule(lr){12-13}
 & ResShift & \textbf{VOSR-0.5B-ms} & \textbf{VOSR-1.4B-ms}
 & StableSR & PASD & SeeSR & DiT4SR
 & SinSR & \textbf{VOSR-0.5B-os} & \textbf{VOSR-1.4B-os}
 & OSEDiff & PiSA-SR \\
\midrule
Params (M)
 & 174.0   & 619.4  & 1{,}742.9
 & 1{,}540 & 1{,}680 & 2{,}510 & 21{,}380
 & 174.0   & 620.7  & 1{,}745.7
 & 1{,}760 & 1{,}290 \\
Time (s)
 & 0.534   & 1.226  & 2.114
 & 10.036  & 2.751  & 3.943  & 12.550
 & 0.116   & 0.094  & 0.095
 & 0.104   & 0.091  \\
\bottomrule
\end{tabular*}
\vspace{-2mm}
\end{table*}

\vspace{-1mm}
\noindent
\textbf{Complexity Comparison}. Table~\ref{tab:complexity} compares model size and inference time of different SR methods with a fixed output resolution of $512 \times 512$ excluding preloading and saving. In the multi-step setting, VOSR is markedly more efficient than T2I-based methods: VOSR-1.4B-ms runs in 2.114s, versus 2.751s for PASD, 3.943s for SeeSR, 10.036s for StableSR, and 12.550s for DiT4SR, while using fewer parameters than most of them. In the one-step setting, both VOSR variants are highly efficient, with VOSR-0.5B-os and VOSR-1.4B-os taking only 0.094s and 0.095s, respectively. They are faster than OSEDiff and comparable to PiSA-SR, while remaining clearly smaller than OSEDiff. These results show that VOSR offers a favorable trade-off between model size and speed in both regimes.

\begin{figure}[t]
    \centering
    \includegraphics[width=0.99\linewidth]{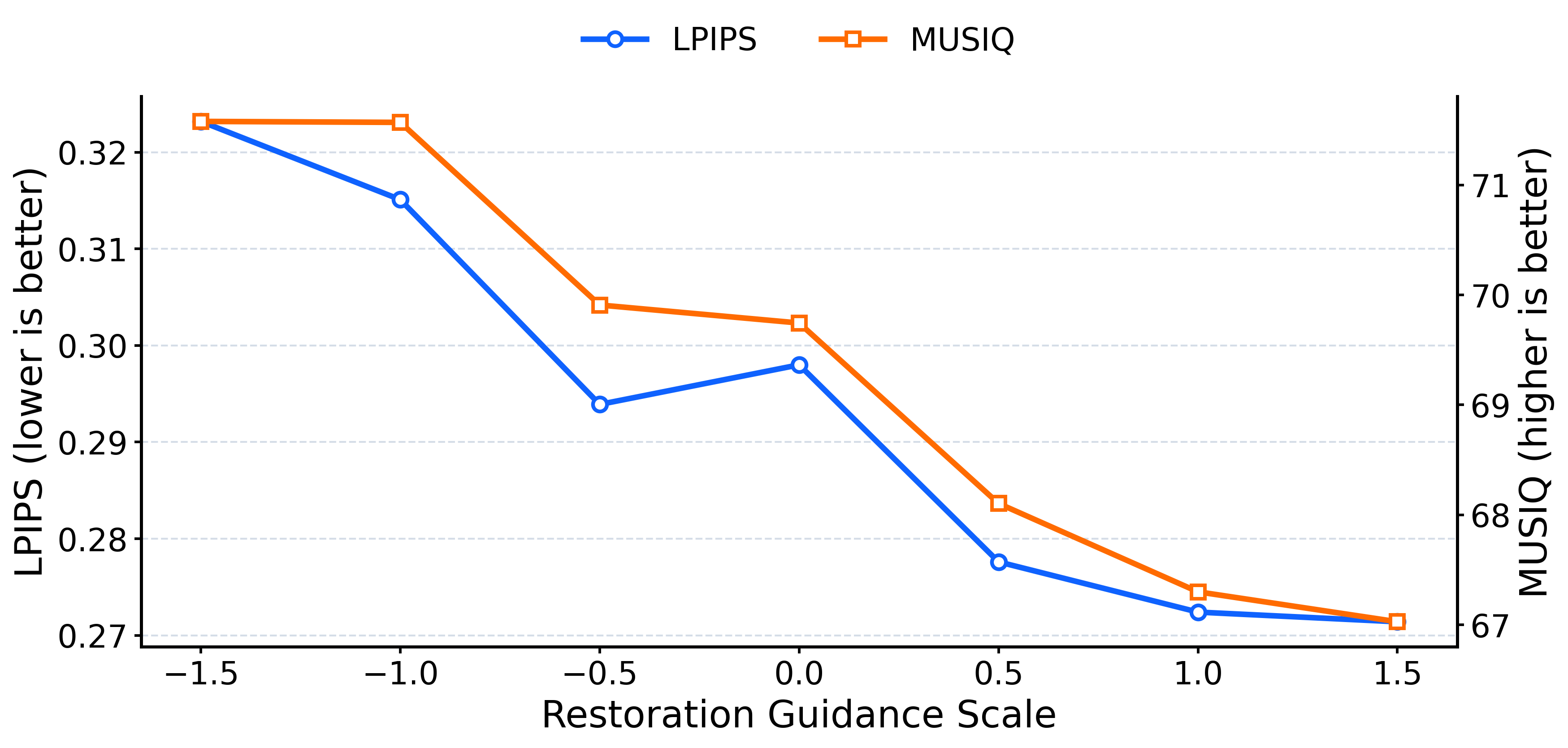}
    \vspace{-2mm}
    \caption{Effect of guidance scale on VOSR-1.4B-ms. As the scale increases, LPIPS rises while MUSIQ drops, indicating a shift from more generative to more faithful restoration to the LR input.}
    \label{fig:guidance-scale}
    \vspace{-5mm}
\end{figure}

\noindent
\textbf{Guidance Scale Behavior}. We further analyze how the guidance scale affects VOSR in Fig.~\ref{fig:guidance-scale}. As the scale increases, LPIPS rises while MUSIQ drops, indicating a shift from more generative results to more faithful results to the LR input. This trend is opposite to that in T2I-based SR \cite{wu2024seesr}, where larger guidance scales usually strengthen injected semantic priors and improve perceptual realism at the cost of faithfulness. The difference comes from the guidance design. In T2I-based methods, CFG mainly amplifies coarse text-aligned semantics on top of a fixed LR control branch. In VOSR, guidance interpolates between a partially conditioned branch and a fully conditioned branch, both still anchored to the LR input. Increasing the guidance scale therefore reduces generative freedom and favors restoration, consistent with our restoration-oriented formulation.

\noindent
\textbf{Additional Studies}. The ablation studies on semantic vision encoder and partial conditioning are in the \textbf{Appendix}.

\section{Conclusion}
In this paper, we revisited generative image SR from a restoration-oriented perspective and showed that high-quality SR can be obtained without T2I pretraining. In particular, we proposed VOSR, a vision-only SR framework grounded in the LR input for generative restoration. By integrating spatially grounded visual semantics with restoration-oriented partial conditioning, VOSR achieved a strong performance of perceptual quality, faithfulness, and efficiency. It demonstrated competitive results with representative T2I-based SR methods on synthetic and real-world benchmarks. In addition, VOSR costed much lower training expenses and supported one-step inference, validating that vision-only generative modeling was a strong alternative to T2I adaptation for real-world SR.

\vspace{+1mm}
\noindent
\textbf{Limitation}.
Our training data scale and model size still lag noticeably behind SR methods built on top of 10B-scale or larger T2I foundation models. We will scale up both the training data and model capacity in future work and extend the framework to broader image restoration tasks.

{
    \small
    \bibliographystyle{ieeenat_fullname}
    \bibliography{main}
}

\clearpage
\appendix
\section{Appendix}

This appendix presents distillation details, ScreenSR benchmark details, training settings, ablation studies, user study results, and additional visual comparisons.

\subsection{Distillation Details}
We follow the notation in Sec.~3.4 of the main paper. Both distilled variants use the same student parameterization $f_{\theta}(z_t,t,r,\kappa)$, where $\kappa$ denotes either the fully conditioned mode $(c_{\mathrm{str}}, c_{\mathrm{sem}})$ or the partially conditioned mode $(\alpha c_{\mathrm{str}}, \varnothing)$. They also share the same restoration-oriented teacher target:
\begin{equation}
v_{\mathrm{tea}}^{\mathrm{guide}} = v_{\mathrm{pcond}} + \omega \left( v_{\mathrm{cond}} - v_{\mathrm{pcond}} \right),
\end{equation}
and the same base objective:
\begin{equation}
\mathcal{L}_{\mathrm{base}} =
\mathbb{E}_{z_0,z_1,t,\kappa}\left[
\left\| f_{\theta}(z_t, t, r{=}t, \kappa) - v_{\mathrm{tea}}^{\mathrm{guide}} \right\|^2
\right].
\end{equation}
Therefore, the semantic condition and the restoration-oriented partial conditioning are kept unchanged during distillation, and the two variants differ only in how they regularize the compressed prediction with $r<t$. In all experiments, the auxiliary consistency loss weight is set to 1.

\subsubsection{Shortcut-based Variant}
We first study a shortcut-based variant adapted from recent shortcut distillation methods~\cite{frans2024one}. Unlike the original formulation, our goal is not to reproduce the full shortcut training pipeline, but to transplant the core idea of self-consistency into our restoration-oriented distillation framework. Specifically, we keep the student parameterization, semantic condition, and partial conditioning design in Sec.~3.4 unchanged, and use a midpoint consistency constraint to regularize the compressed prediction from $t$ to $r$.

For a sampled compressed target time $r<t$, we define the midpoint $s=(t+r)/2$. Let
$u_{t\rightarrow r}=f_{\theta}(z_t,t,r,\kappa)$ and 
$u_{t\rightarrow s}=f_{\theta}(z_t,t,s,\kappa)$.
We then construct an intermediate latent using the detached first-half prediction:
\begin{equation}
z_s = z_t - (t-s)\,\mathrm{sg}(u_{t\rightarrow s}),
\end{equation}
and predict the second half as:
\begin{equation}
u_{s\rightarrow r}=f_{\theta}(z_s,s,r,\kappa).
\end{equation}
We use the detached two-stage decomposition to regularize the direct compressed prediction:
\begin{equation}
\bar u_{t\rightarrow r}
=
\mathrm{sg}\left(\frac{u_{t\rightarrow s}+u_{s\rightarrow r}}{2}\right),
\end{equation}
\begin{equation}
\mathcal{L}_{\mathrm{short\text{-}cons}}
=
\left\|
u_{t\rightarrow r}-\bar u_{t\rightarrow r}
\right\|_2^2.
\end{equation}
The total objective of this variant is:
\begin{equation}
\mathcal{L}_{\mathrm{short}}
=
\mathcal{L}_{\mathrm{base}}
+
\mathcal{L}_{\mathrm{short\text{-}cons}}.
\end{equation}
This design can be viewed as a shortcut-inspired consistency regularizer tailored to our SR setting, rather than a strict reproduction of the original shortcut method.

\subsubsection{Recursive-Consistency-based Variant}
We further adapt the recursive-consistency (RC) based strategy \cite{sun2026anystep} under the same conditioning and teacher-guidance setup. For a compressed prediction from $t$ to $r$, we first compute
\begin{equation}
u_{t\rightarrow r}=f_{\theta}(z_t,t,r,\kappa).
\end{equation}
We then perform a teacher-guided warm start from $z_t$ to an intermediate time $t_m=\max(t-\Delta t,r)$:
\begin{equation}
z_{t_m}=z_t-(t-t_m)\,v_{\mathrm{tea}}^{\mathrm{guide}}.
\end{equation}
Starting from $z_{t_m}$, we run a detached multi-step ODE rollout to time $r$ under the same conditioning mode $\kappa$, which yields a recursive trajectory target, denoted by $u_{t_m\rightarrow r}^{\mathrm{tar}}$. Following the RC formulation, we construct a corrected detached target as:
\begin{equation}
\mathrm{corr}
=
\mathrm{clip}\left(
c_l\,u_{t\rightarrow r}
-
c_r\,u_{t_m\rightarrow r}^{\mathrm{tar}}
-
v_{\mathrm{tea}}^{\mathrm{guide}},
[-1,1]
\right),
\end{equation}
\begin{equation}
\tilde u_{t\rightarrow r}
=
\mathrm{sg}(u_{t\rightarrow r})-\mathrm{corr},
\end{equation}
and optimize
\begin{equation}
\mathcal{L}_{\mathrm{rc\text{-}cons}}
=
\left\|
u_{t\rightarrow r}-\tilde u_{t\rightarrow r}
\right\|_2^2.
\end{equation}
The resulting objective is:
\begin{equation}
\mathcal{L}_{\mathrm{rc}}
=
\mathcal{L}_{\mathrm{base}}
+
\mathcal{L}_{\mathrm{rc\text{-}cons}}.
\end{equation}

Compared with shortcut-based regularization, the advantage of RC-based regularization is that it explicitly pulls the student trajectory toward the teacher trajectory, rather than matching a target induced by randomly sampled intermediate points. This makes the supervision more aligned with the teacher's denoising path under large temporal compression. We therefore adopt the RC variant in the main paper. 

To quantitatively compare the two distilled variants, we further report their performance on RealSR using VOSR-0.5B-ms as the teacher. As shown in Table~\ref{tab:distill_compare}, both shortcut-based and RC-based distillations preserve strong perceptual performance after compression to one-step inference. Compared with the multi-step teacher, shortcut distillation improves LPIPS while maintaining competitive MUSIQ, indicating that the distilled student can inherit the teacher's perceptual restoration capability. RC-based distillation performs better overall, achieving both lower LPIPS and higher MUSIQ than the shortcut-based variant. This result is consistent with our observation that RC provides a stronger training signal under large temporal compression and is thus better suited to one-step generative SR.

\begin{table}[t]
\centering
\caption{Comparison of shortcut-based and RC-based distillation on RealSR using VOSR-0.5B-ms as the teacher.}
\label{tab:distill_compare}

\scriptsize
\setlength{\tabcolsep}{0pt}
\renewcommand{\arraystretch}{1.1}
\begin{tabular*}{\columnwidth}{@{\extracolsep{\fill}}lcc@{}}
\toprule
\textbf{Method} & \textbf{LPIPS}$\downarrow$ & \textbf{MUSIQ}$\uparrow$ \\
\midrule
Teacher (VOSR-0.5B-ms) & 0.3069 & 68.93 \\
Shortcut-based distillation & 0.2913 & 68.21 \\
RC-based distillation & 0.2856 & 69.78 \\
\bottomrule
\end{tabular*}
\vspace{-2mm}
\end{table}

\subsection{ScreenSR Benchmark Details}

\begin{figure*}[t]
  \centering
  \includegraphics[width=\linewidth]{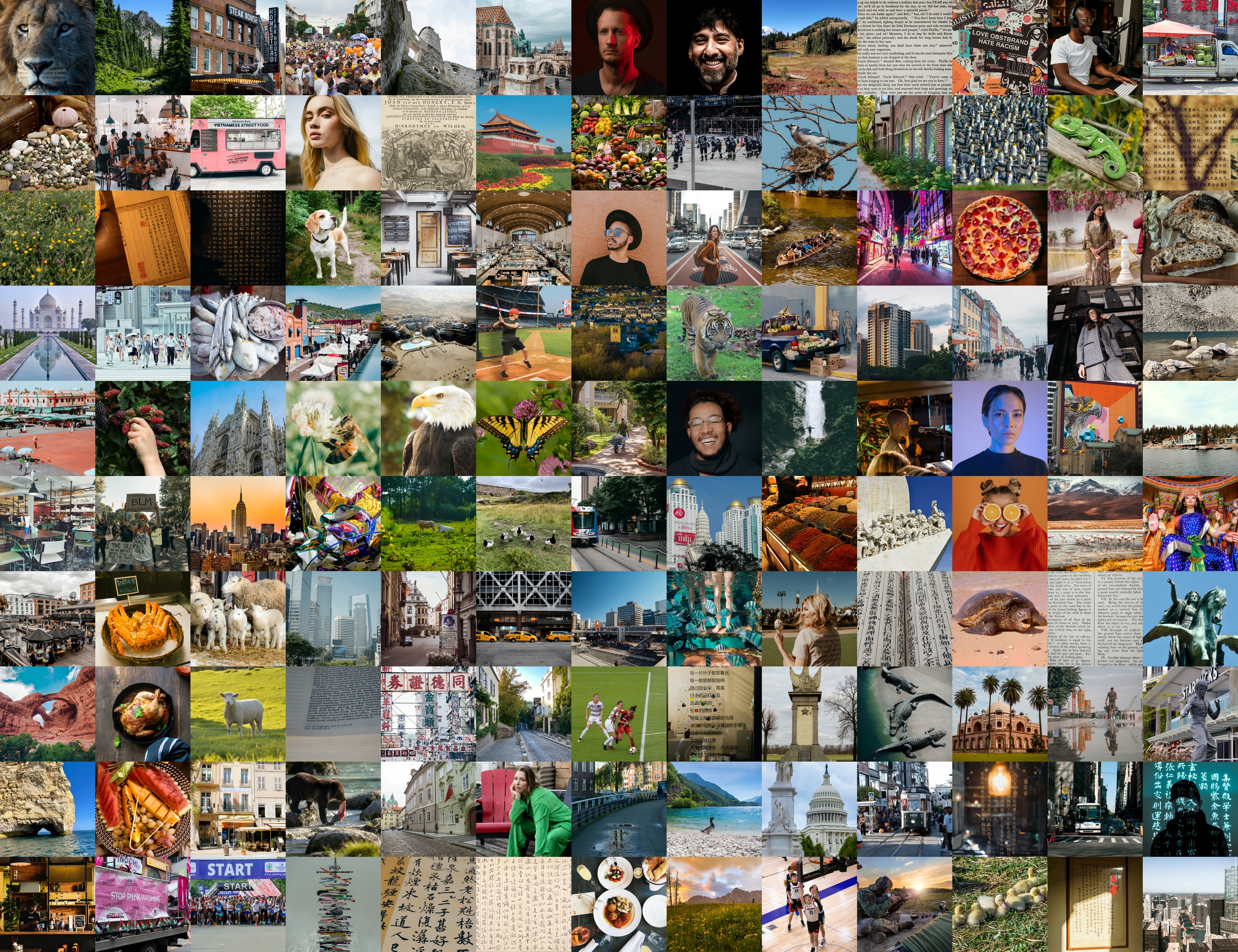}
  \vspace{-3mm}
  \caption{Thumbnail montage of the ScreenSR benchmark. The selected 130 GT images cover diverse  scenarios, including indoor and outdoor scenes, humans, animals, plants, artworks, and multilingual text, with substantial variation in object and scene scales. This diversity ensures a comprehensive evaluation of generative SR methods in terms of semantic coverage, structural fidelity, and robustness across different content types.}
  \label{fig:screensr-montage}
  \vspace{-4mm}
\end{figure*}

ScreenSR is designed as a real-world paired benchmark for evaluating generative SR in practical mobile-photography scenarios. We first manually collect a diverse set of high-quality source images from the web and deliberately curate them with balanced scene categories and object/scene scales. The selected images cover indoor scenes, outdoor scenes, human subjects, animals, plants, artworks, and Chinese and English text. We also include static and dynamic scenes and intentionally keep examples at different spatial scales within each major category, so that the benchmark can better evaluate SR methods from various aspects, including semantic diversity, structural fidelity, and robustness across scale variations. A thumbnail montage of the selected 130 images is shown in Fig.~\ref{fig:screensr-montage}.

After content curation, the source images are displayed on a high-resolution screen and re-photographed by flagship smartphones to construct paired real-world LR-HR examples. The data capturing devices cover flagship models from major smartphone manufacturers, including OPPO, vivo, Xiaomi, and Huawei. To ensure reliable pairing, each source image is first placed on a white canvas with four ArUco markers near the border. After capture, the detected marker corners are used to estimate a geometric transform that warps the mobile photo back to the original image plane, producing pixel-aligned pairs at the target resolution. We further apply a wavelet-based low-frequency color alignment while preserving high-frequency details from the captured image. This screen re-photography pipeline preserves accurate pairing while introducing realistic mobile imaging degradations. Compared with purely synthetic degradations, the resulting LR images better reflect practical mobile captures, while the displayed source images provide clean and visually strong references. The final ScreenSR benchmark contains 130 paired samples, all used for zero-shot real-world evaluation in our experiments.

A key motivation for building ScreenSR is that the quality of GT images matters for real-world SR evaluation. If the GT itself has limited perceptual quality, the reliability of benchmark conclusions may be weakened. We compare the no-reference quality metrics of GT images from ScreenSR, RealSR, and DRealSR. As shown in Table~\ref{tab:gt_quality_benchmark}, ScreenSR consistently achieves substantially better GT quality on all five no-reference metrics, including NIQE, MUSIQ, MANIQA, AFINE-NR, and TOPIQ-NR. These results support our claim that ScreenSR provides cleaner and more reliable references for real-world paired evaluation.

\begin{table}[t]
\centering
\caption{No-reference quality comparison of GT images in different real-world paired SR benchmarks. Better GT quality is indicated by lower NIQE and AFINE-NR, and higher MUSIQ, MANIQA, and TOPIQ-NR.}
\label{tab:gt_quality_benchmark}
\vspace{-2mm}
\fontsize{6.5pt}{7.5pt}\selectfont
\setlength{\tabcolsep}{3pt}
\begin{tabular*}{\columnwidth}{@{\extracolsep{\fill}}lccccc@{}}
\toprule
\textbf{Benchmark} & \textbf{NIQE}$\downarrow$ & \textbf{MUSIQ}$\uparrow$ & \textbf{MANIQA}$\uparrow$ & \textbf{AFINE-NR}$\downarrow$ & \textbf{TOPIQ-NR}$\uparrow$ \\
\midrule
ScreenSR & \textbf{3.7719} & \textbf{72.1500} & \textbf{0.7187} & \textbf{-1.2093} & \textbf{0.7363} \\
RealSR \cite{realsr}   & 6.1167 & 57.4564 & 0.6016 & -0.9088 & 0.4140 \\
DRealSR \cite{drealsr}  & 6.7909 & 50.5644 & 0.5588 & -0.7731 & 0.3932 \\
\bottomrule
\end{tabular*}
\vspace{-4mm}
\end{table}

\subsection{Detailed Training Settings}
We train VOSR in a progressive manner. Specifically, the multi-step model is first pretrained at $256\times256$ resolution for 400K steps with a global batch size of 1024 and a constant learning rate of $1.0\times10^{-4}$, and is then further trained at $512\times512$ resolution for another 400K steps with a global batch size of 256 and a constant learning rate of $5.0\times10^{-5}$. After obtaining the multi-step teacher, we distill it into a one-step model for 50K steps using a batch size of 32 and a constant learning rate of $2.0\times10^{-5}$. Across all stages, we use no warm-up, set the weight decay to 0.01, the gradient clipping threshold to 1.0, and the EMA decay to 0.9999, and adopt AdamW optimizer with $\beta_1=0.9$ and $\beta_2=0.95$. For the diffusion backbone, both VOSR-0.5B and VOSR-1.4B use a patch size of 2 and an MLP ratio of 4. VOSR-0.5B uses dimension 1024, depth 28, and 16 attention heads, while VOSR-1.4B uses dimension 1536, depth 36, and 24 attention heads.
\begin{table}[t]
\scriptsize
\centering
\caption{Training recipe for VOSR, including the progressive pretraining for the multi-step models and the distillation for the one-step model.}
\label{tab:vosr_progressive_recipe}
\resizebox{0.98\linewidth}{!}{
\begin{tabular}{lccc}
\toprule
 & PT $256_{px}$ & PT $512_{px}$ & Distill to One-step \\
\midrule
Learning rate & $1.0\times10^{-4}$ & $5.0\times10^{-5}$ & $2.0\times10^{-5}$ \\
LR scheduler & Constant & Constant & Constant \\
Warm-up steps & 0 & 0 & 0 \\
Training steps & 400K & 400K & 50K \\
Global batch size & 1024 & 256 & 32 \\
Weight decay & 0.01 & 0.01 & 0.01 \\
Gradient clip & 1.0 & 1.0 & 1.0 \\
EMA & 0.9999 & 0.9999 & 0.9999 \\
Optimizer & \multicolumn{3}{c}{AdamW ($\beta_1=0.9$, $\beta_2=0.95$)} \\
\bottomrule
\end{tabular}
}
\end{table}

\subsection{Ablation Studies}
Unless otherwise specified, all ablation experiments are conducted on VOSR-0.5B, and trained at $512\times512$ resolution for 100K steps.

\subsubsection{Effect of Visual Semantic Condition}

Table~\ref{tab:ablation_semantic} evaluates the visual semantic condition on RealSR. Removing the semantic condition causes a clear degradation, showing that structural conditioning alone is insufficient for challenging real-world SR. Adding pretrained visual semantic features consistently improves perceptual quality across different encoders. CLIP achieves the best LPIPS, while DINO-based encoders produce notably higher MUSIQ, indicating better perceptual realism. We use DINOv2 in the final model as a balanced choice considering both overall performance and stability.

\subsubsection{Partial Conditioning}

We compare three guidance designs on LSDIR: using the fully conditioned model alone without guidance, standard CFG with a fully unconditional auxiliary branch, and our restoration-oriented partial conditioning. For our method, instead of fixing the structural retention factor to a single value, we randomly sample $\alpha$ within $[0.05, 0.25]$ during training. We adopt this design because the partially conditioned branch is intended to represent a family of weakly conditioned, input-anchored restoration states rather than one specific conditioning strength. Randomizing $\alpha$ exposes the model to diverse weak-structure conditions, which improves the robustness of the auxiliary branch and avoids overfitting the guidance behavior to a narrow partial-conditioning regime. The sampled range is kept small so that the partial branch remains weaker than the fully conditioned one while still preserving minimal structural anchors from the LR input.

Table~\ref{tab:ablation_partial} shows that explicit guidance is important for high-quality generative SR. Compared with using the fully conditioned model alone, standard CFG leads to a clear performance degradation. This is due to the fact that, under limited training data and computation, the fully unconditional branch is difficult to train from scratch for SR with standard CFG. Since the unconditional branch is poorly learned, the resulting guidance becomes unstable and can even harm restoration quality. In contrast, our partial conditioning achieves the best results, suggesting that an input-anchored auxiliary branch is much easier to optimize and provides a more reliable guidance direction for balancing perceptual realism and restoration fidelity.

\begin{table}[t]
\centering
\caption{Ablation on visual semantic encoders on RealSR.}
\label{tab:ablation_semantic}
\scriptsize
\setlength{\tabcolsep}{0pt}
\renewcommand{\arraystretch}{1.1}
\begin{tabular*}{\columnwidth}{@{\extracolsep{\fill}}lcc@{}}
\toprule
\textbf{Method} & \textbf{LPIPS}$\downarrow$ & \textbf{MUSIQ}$\uparrow$ \\
\midrule
w/o SVE           & 0.3011 & 63.74 \\
w/ SVE (CLIP)     & 0.2788 & 63.82 \\
w/ SVE (SigLIPv2) & 0.2817 & 64.81 \\
w/ SVE (DINOv3)   & 0.2858 & 67.51 \\
w/ SVE (DINOv2)   & 0.2872 & 68.23 \\
\bottomrule
\end{tabular*}
\vspace{-4mm}
\end{table}

\begin{figure}[t]
  \centering
  \includegraphics[width=1\linewidth]{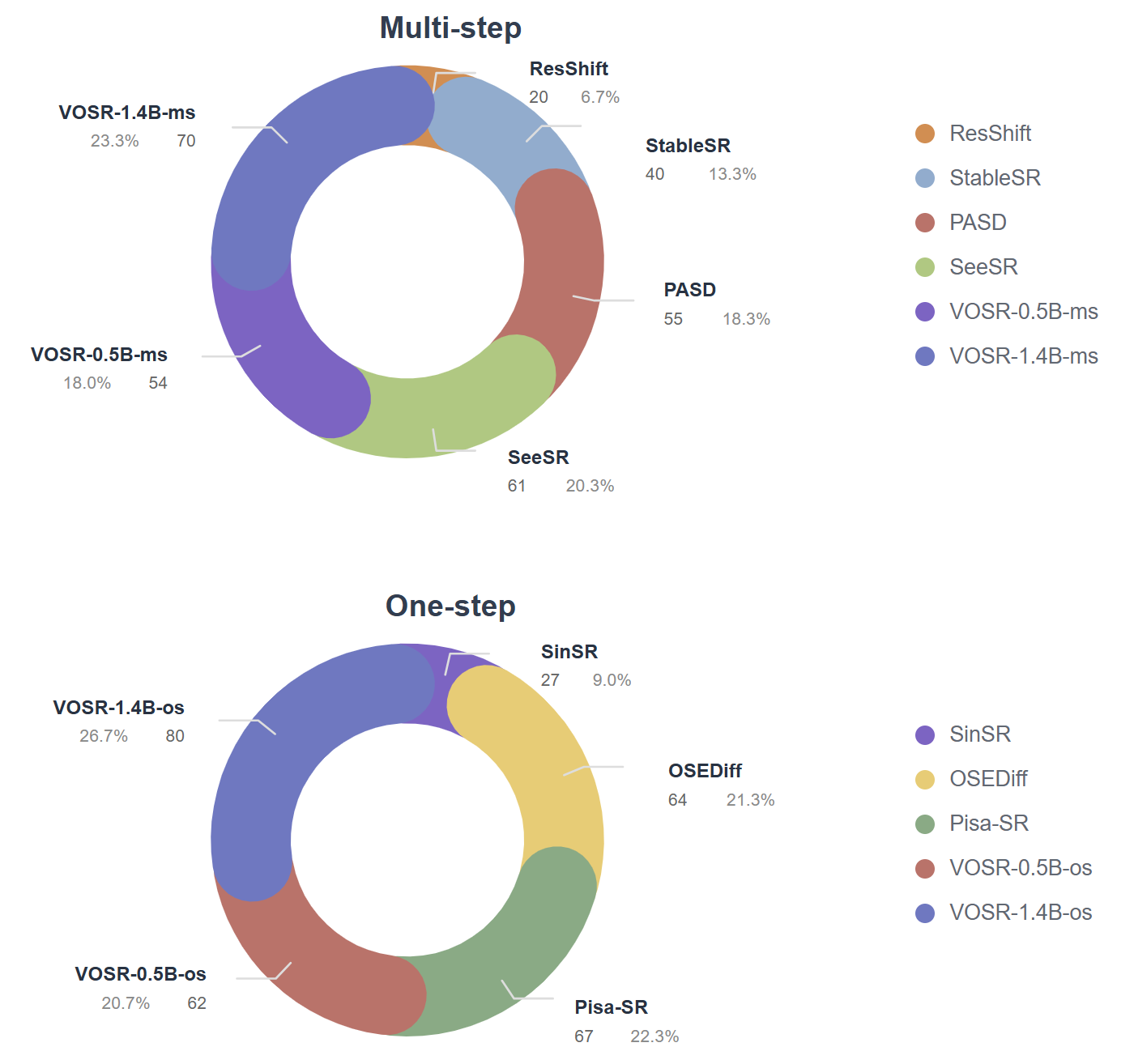}
  \caption{
    User study results in the multi-step and one-step settings. VOSR-1.4B-ms and VOSR-1.4B-os receive the highest numbers of votes in their respective groups, showing strong human preference in perceptual quality and consistency with the LR input.
  }
  \label{fig:user_study}
\end{figure}

\begin{table}[t]
\centering
\caption{Ablation studies on restoration-oriented partial conditioning on LSDIR.}
\label{tab:ablation_partial}
\scriptsize
\setlength{\tabcolsep}{0pt}
\renewcommand{\arraystretch}{1.1}
\begin{tabular*}{\columnwidth}{@{\extracolsep{\fill}}lcc@{}}
\toprule
\textbf{Guidance design} & \textbf{LPIPS}$\downarrow$ & \textbf{MUSIQ}$\uparrow$ \\
\midrule
Full condition only & 0.3752 & 67.29 \\
Standard CFG & 0.4053 & 50.78 \\
Ours (partial conditioning) & 0.3772 & 69.26 \\
\bottomrule
\end{tabular*}
\vspace{-2mm}
\end{table}

\begin{figure*}[t]
  \centering
  \includegraphics[width=\linewidth]{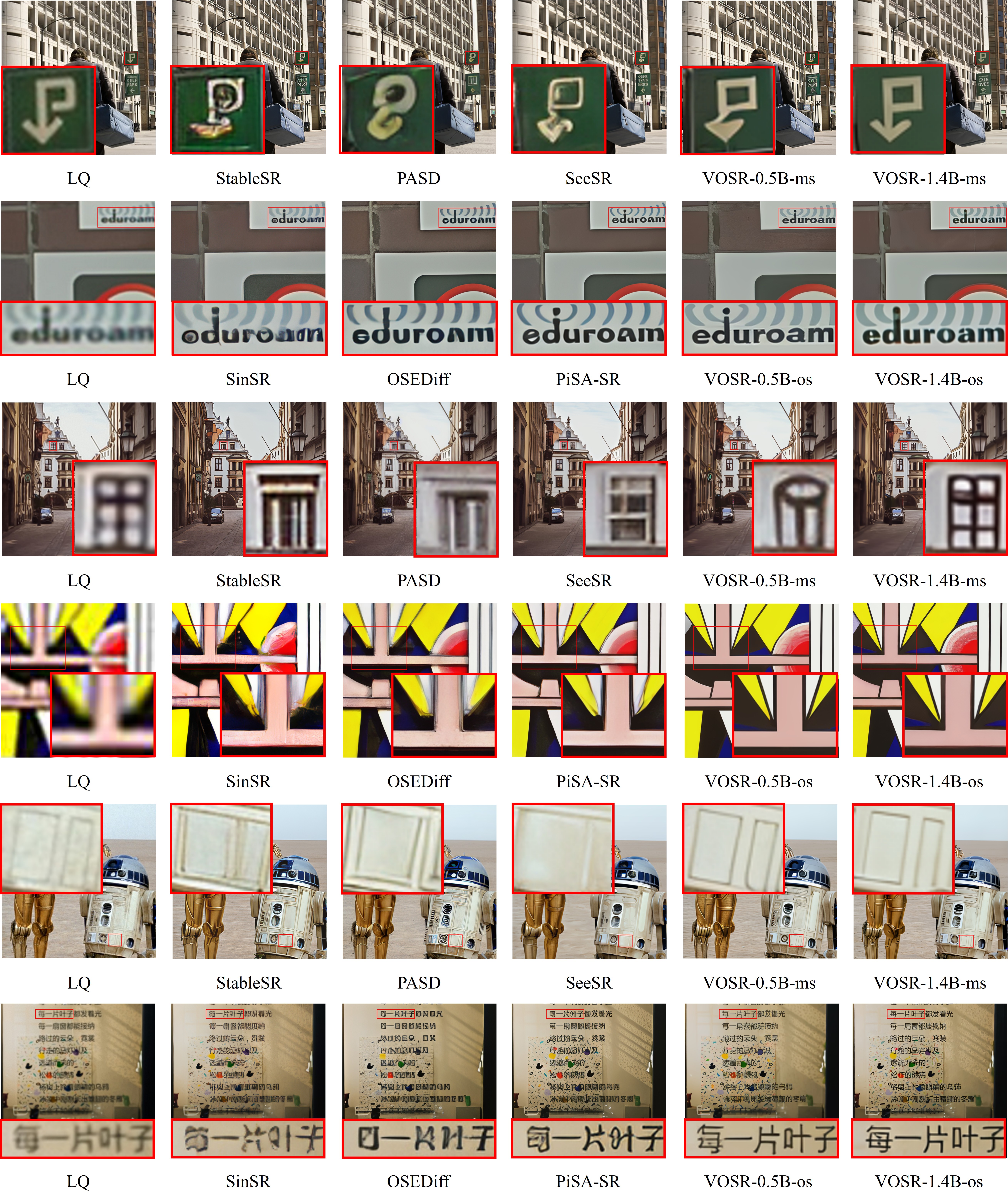}
  \vspace{-4mm}
  \caption{Additional visual comparisons of multi-step (1st, 3rd and 5th) and one-step (2nd, 4th and 6th) SR results. Compared with representative vision-only and T2I-based methods, VOSR produces perceptually more realistic details while better preserving structures that are faithful to the LR input. Please zoom in for better view.}
  \label{fig:vis-comp-suppl}
\end{figure*}

\subsection{User Study}

To further validate VOSR from a human perceptual perspective, we conduct separate user studies for the multi-step and one-step settings using 30 LR images. For each sample, participants are shown the LR input and the SR results from all compared methods, and they are asked to select the best result. The evaluation uses two equally weighted criteria: (1) perceptual quality and (2) consistency with the LR input, including structural and texture fidelity. Ten volunteers were invited and each evaluated all samples.

In the multi-step setting, we compare ResShift \cite{yue2023resshift}, StableSR \cite{wang2023exploiting}, PASD \cite{yang2023pixel}, SeeSR \cite{wu2024seesr}, VOSR-0.5B-ms, and VOSR-1.4B-ms (300 votes total). As shown in Fig.~\ref{fig:user_study}, VOSR-1.4B-ms wins most votes (70/300, 23.3\%), followed by SeeSR (61/300, 20.3\%), PASD (55/300, 18.3\%), and VOSR-0.5B-ms (54/300, 18.0\%); StableSR and ResShift receive 40/300 (13.3\%) and 20/300 (6.7\%). For the one-step setting, we compare SinSR \cite{wang2023sinsr}, OSEDiff \cite{wu2024one}, PiSA-SR \cite{sun2024pixel}, VOSR-0.5B-os, and VOSR-1.4B-os, resulting in 300 total votes. VOSR-1.4B-os ranks first with 80/300 votes (26.7\%), followed by PiSA-SR (67/300, 22.3\%), OSEDiff (64/300, 21.3\%), and VOSR-0.5B-os (62/300, 20.7\%), while SinSR receives 27/300 votes (9.0\%). These results show that VOSR is strongly preferred by human evaluators in both multi-step and one-step settings, validating its ability to achieve better perceptual quality while preserving stronger input faithfulness.

\subsection{More Visual Results}

Fig.~\ref{fig:vis-comp-suppl} provides six additional visual comparisons, including three multi-step cases (1st, 3rd and 5th) and three one-step cases (2nd, 4th and 6th). Overall, T2I-based methods often produce visually sharp results, but are less reliable in recovering fine structures that need to be faithful to the LR input. In contrast, VOSR consistently restores more local details with clearer structure and fewer hallucinations.

In the first example, the sign contains a thin symbol contour and a sharp corner structure that are difficult to recover from the degraded input. PASD \cite{yang2023pixel} and SeeSR \cite{wu2024seesr} produce obvious shape distortions, while StableSR \cite{wang2023exploiting} restores the symbol more plausibly but with inaccurate local geometry. By contrast, the VOSR variants recover clearer and more faithful symbol shapes, with VOSR-1.4B-ms producing the most complete contour and corner details. Similar trends can be observed in other multi-step examples: VOSR preserves sharper window boundaries and clearer panel edges, while T2I-based methods either blur the structures or generate less accurate local geometry. Meanwhile, compared with these T2I-based methods, VOSR runs substantially faster, as shown by the complexity comparison in the main paper.

For the one-step examples, similar conclusions can be drawn. In text and symbol regions, SinSR shows weaker generative ability, while OSEDiff and PiSA-SR sometimes produce sharper, yet less faithful shapes. By contrast, VOSR restores clearer characters, cleaner boundaries, and more stable local structures. In particular, the two VOSR one-step models recover the sign content and thin edges more faithfully while remaining highly efficient.

These visual results are consistent with the quantitative comparisons and efficiency analysis in the main paper. They further support that VOSR benefits from a restoration-oriented design: by combining spatially grounded visual semantics with input-anchored guidance, it can generate perceptually strong details without relying on the generative priors transferred from pre-trained T2I models.

\end{document}